\documentclass{article}
\usepackage{arxiv}
\usepackage[utf8]{inputenc} 
\usepackage[T1]{fontenc}    
\usepackage[hyperfootnotes=false]{hyperref}       
\usepackage{url}            
\usepackage{booktabs}       
\usepackage{amsfonts}       
\usepackage{nicefrac}       
\usepackage{microtype}      
\usepackage{lipsum}		
\usepackage{graphicx}
\usepackage{grffile}
\usepackage[numbers]{natbib}
\usepackage{doi}
\usepackage{amsmath}
\usepackage{amsthm}
\usepackage{subcaption}
\usepackage{footnote}
\makesavenoteenv{table*}
\makesavenoteenv{tabular}
\usepackage{tablefootnote}
\usepackage{mwe}
\usepackage{float}
\usepackage{ragged2e}
\usepackage{multirow}
\usepackage{wrapfig}
\usepackage{xcolor}
\usepackage{setspace}
\floatstyle{plaintop}
\restylefloat{table}

\usepackage{etoolbox}

\title{VIVAT: Virtuous Improving VAE Training \\through Artifact Mitigation}

\author{
\hspace{1cm}\textbf{Lev Novitskiy}$^{1,*}$,\hspace{1cm} \And \hspace{1cm}\textbf{Viacheslav Vasilev}$^{1}$,\hspace{1cm} \And \hspace{1cm}\textbf{Maria Kovaleva}$^{1}$,\hspace{1cm} 
\And \textbf{Vladimir Arkhipkin}$^1$, \And \textbf{Denis Dimitrov}$^{1}$
}

\renewcommand{\shorttitle}{VIVAT: Virtuous Improving VAE Training through Artifact Mitigation}

\begin{document}

\maketitle

\begin{abstract}
Variational Autoencoders (VAEs) remain a cornerstone of generative computer vision, yet their training is often plagued by artifacts that degrade reconstruction and generation quality. This paper introduces VIVAT, a systematic approach to mitigating common artifacts in KL-VAE training without requiring radical architectural changes. We present a detailed taxonomy of five prevalent artifacts -- color shift, grid patterns, blur, corner and droplet artifacts -- and analyze their root causes. Through straightforward modifications, including adjustments to loss weights, padding strategies, and the integration of Spatially Conditional Normalization, we demonstrate significant improvements in VAE performance. Our method achieves state-of-the-art results in image reconstruction metrics (PSNR and SSIM) across multiple benchmarks and enhances text-to-image generation quality, as evidenced by superior CLIP scores. By preserving the simplicity of the KL-VAE framework while addressing its practical challenges, VIVAT offers actionable insights for researchers and practitioners aiming to optimize VAE training.
\end{abstract}

\section{Introduction}

Variational Autoencoders (VAEs) \cite{Kingma2013AutoEncodingVB, pmlr-v32-rezende14} have long been established as a cornerstone of generative AI, playing a pivotal role in tasks such as image reconstruction and image generation \cite{10.1007/978-3-319-46493-0_47, 10.5555/3454287.3455618, vahdat2020NVAE, esser2020taming, yu2022vectorquantizedimagemodelingimproved}. Their significance is further emphasized by their integration into modern latent diffusion pipelines \cite{Rombach2021HighResolutionIS, chen2023pixartalpha, esser2024scalingrectifiedflowtransformers, flux2024, podell2023sdxlimprovinglatentdiffusion, arkhipkin2024kandinsky30technicalreport, vladimir-etal-2024-kandinsky, chen2024deep}, where they serve as the main ingredient for latent space formation. Despite their widespread adoption and extensive research history, the training of high-quality VAEs remains fraught with challenges, particularly with respect to the emergence of artifacts that degrade generation quality \cite{kouzelis2025eqvaeequivarianceregularizedlatent, skorokhodov2025improvingdiffusabilityautoencoders}. These persistent issues underscore the continued relevance of refining VAE training methodologies to enhance their robustness and generative fidelity.

Previous research has explored various architectural modifications and different concepts of VAEs, leading to the development of specialized model families and various changes in the optimization procedure \cite{Higgins2016betaVAELB, 10.5555/3295222.3295378, 10.5555/3454287.3455618, pmlr-v139-rybkin21a, vahdat2020NVAE, child2021deepvaesgeneralizeautoregressive, jiang2021focal, pandey2022diffusevaeefficientcontrollablehighfidelity, chen2024deep, kouzelis2025eqvaeequivarianceregularizedlatent}. Although these contributions have expanded the theoretical understanding of VAEs, many works focus on fundamental innovations while neglecting practical considerations, particularly the systematic analysis and mitigation of training artifacts. Consequently, despite the diversity of proposed solutions, there remains a lack of straightforward and actionable strategies to address anomalies in existing VAE implementations and training process. In this work, we conduct a comprehensive classification and analysis of these artifacts, offering practical, easy-to-implement recommendations to eliminate them without resorting to radical architectural overhauls.

We posit that the potential of conventional autoencoding frameworks remains underutilized and that significant improvements can be achieved through targeted refinements rather than entirely new methodologies. To this end, we adopt the classic KL-VAE~\footnote{Later in this paper, the terms ``VAE'' and ``KL-VAE'' will be used interchangeably, unless otherwise stated.} architecture \cite{Kingma2013AutoEncodingVB} -- a proven and widely used framework that has been successfully applied in prior work for modern generative models \cite{chen2023pixartalpha, flux2024, esser2024scalingrectifiedflowtransformers}. By systematically investigating the origins of training artifacts within this architecture, we propose simple yet effective solutions that yield measurable improvements in model performance. Quantitative evaluations demonstrate that our approach achieves state-of-the-art results for reconstruction quality without introducing unnecessary complexity or departing from established VAE training principles. With this work, we aim to provide valuable insights for researchers and practitioners who are seeking to optimize VAE training while preserving the simplicity and efficiency of classical autoencoding methods.

Thus, the key contributions of this work are as follows:

\begin{itemize}
    \item We provide a detailed taxonomy of five common artifacts in KL-VAE training, identifying their potential root causes and their impact on reconstruction quality;
    \item We propose straightforward, easy-to-implement modifications to mitigate these artifacts, ensuring a stable training process without requiring fundamental architectural changes;
    \item We demonstrate that our approach enhances VAE performance, achieving superior results in reconstruction quality while maintaining the simplicity of the KL-VAE framework;
    \item Finally, we show that VAE, trained according to our methodology, is well-suited for the text-to-image generation task and demonstrates a high-level quality that surpasses the previous state-of-the-art image reconstruction model, namely Flux VAE \cite{flux2024}, in terms of CLIP score.
\end{itemize}

This paper is organized as follows: In Section \ref{sec:vae_theory}, we provide a brief overview of classic KL-VAE and the training losses used in our work. We then describe the basic KL-VAE architecture without any improvements in Section \ref{sec:vae_arch}. Finally, we briefly mention the role of VAE in the image generation pipeline in Section \ref{sec:ldm}. Section \ref{sec:challenges} focuses on common issues that arise during KL-VAE training and the reasons behind them. Section \ref{sec:methods} presents our proposed solutions to these problems, while Section \ref{sec:results} presents the results of our improved method for image reconstruction and text-to-image generation.

\section{Related works}

\subsection{VAE and its modifications}

The classical VAE, also referred to as KL-VAE, was introduced by \cite{Kingma2013AutoEncodingVB} and quickly adopted for conditional image generation \cite{10.1007/978-3-319-46493-0_47}. The core principle of this approach involves minimizing the Kullback-Leibler (KL) divergence between the latent distribution and a Gaussian prior. Since its inception, this architecture has been widely utilized and extensively modified. Notable architectural innovations include hierarchical VAE \cite{vahdat2020NVAE} and DiffuseVAE \cite{pandey2022diffusevaeefficientcontrollablehighfidelity}, the latter employing diffusion processes to refine the blurry VAE outputs. Previous research has also explored the impact of architectural depth in autoencoders \cite{child2021deepvaesgeneralizeautoregressive} and proposed a VAE variant optimized for high compression ratios (up to 128$\times$) \cite{chen2024deep}. However, the majority of prior work has focused on developing novel loss functions and regularization techniques, such as spectral regularization \cite{vahdat2020NVAE}, frequency-domain optimization \cite{jiang2021focal}, the integration of Wasserstein distance \cite{tolstikhin2019wassersteinautoencoders}, and more \cite{Higgins2016betaVAELB, pmlr-v139-rybkin21a, kouzelis2025eqvaeequivarianceregularizedlatent, skorokhodov2025improvingdiffusabilityautoencoders, dilokthanakul2017deepunsupervisedclusteringgaussian, 10.1609/aaai.v33i01.33015066, 10.5555/3540261.3541252, pmlr-v80-zhao18b}. Despite these advancements, most of these contributions have not been widely adopted in large-scale commercial generative models. An alternative approach, VQ-VAE \cite{10.5555/3295222.3295378}, replaces the continuous latent space with a discrete codebook, leading to significant improvements in generative tasks \cite{10.5555/3454287.3455618, esser2020taming, yu2022vectorquantizedimagemodelingimproved, arkhipkin2024kandinsky30technicalreport}. This method has inspired numerous subsequent modifications \cite{10.5555/3454287.3455618, zheng2022movq}, including VQ-GAN \cite{esser2020taming}, which enhances VQ-VAE through adversarial and perceptual losses. Nevertheless, quantized VAEs are known to produce geometric artifacts, and scaling their latent space dimensions is non-trivial. As a result, KL-VAE remains prevalent in several recent high-profile generative models \cite{chen2023pixartalpha, esser2024scalingrectifiedflowtransformers, flux2024}. In this work, we further investigate the KL-VAE architecture, aiming to advance its practical potential in reconstruction and generative applications.

\subsection{VAE for latent diffusion}

Due to inherent limitations in sample quality, VAEs are not the best option for generation tasks on their own. However, they have become a prominent component in latent diffusion models \cite{Rombach2021HighResolutionIS}. By reducing the spatial dimensions of images, VAEs enable more efficient representations, replacing computationally expensive pixel-based approaches. Latent diffusion models for image generation employ both continuous KL-VAEs \cite{chen2023pixartalpha, esser2024scalingrectifiedflowtransformers, flux2024} and quantized autoencoders \cite{ramesh2022hierarchicaltextconditionalimagegeneration, arkhipkin2024kandinsky30technicalreport}. Key principles include increasing the channel size of latent variables and spatial compression ratios. When extending diffusion pipelines to video generation, modified VAEs were developed to incorporate temporal dependencies, either by preserving channel dimensions \cite{blattmann2023videoldm, 10815947} or compressing along the temporal axis \cite{yang2025cogvideoxtexttovideodiffusionmodels, kong2025hunyuanvideosystematicframeworklarge}. Since video VAEs build upon the image VAEs, our work focuses on analyzing spatial artifacts in image reconstruction, without loss of generality for the proposed solutions.

\subsection{VAE artifacts}

The problem of artifacts in image reconstruction using VAEs has been widely studied. A well-known limitation of VAEs is their tendency to produce blurred outputs, attributed to the asymmetric nature of the KL divergence \cite{Kingma_2019}. While adversarial loss can mitigate blurring, it often introduces visual artifacts \cite{Larsen2015AutoencodingBP} and training instability, particularly at high resolutions \cite{chen2024deep}. Additionally, VAEs are prone to losing high-frequency information \cite{jiang2021focal}. Stronger regularization (e.g., increasing the KL-divergence weight) yields a smoother latent space but reduces its information capacity, leading to loss of fine details and degraded reconstruction quality \cite{kouzelis2025eqvaeequivarianceregularizedlatent, 10.1007/978-3-031-72998-0_17}. Numerous studies have proposed improved training methods to reduce artifacts, yet most lack a systematic analysis or taxonomy of the specific visual artifacts encountered \cite{pmlr-v139-rybkin21a, esser2020taming, jiang2021focal, pandey2022diffusevaeefficientcontrollablehighfidelity}. \cite{kouzelis2025eqvaeequivarianceregularizedlatent} highlights a connection between artifacts and insufficient equivariance in the latent space under spatial transformations, though without explicitly categorizing the artifacts. Recently, visual artifacts in VAEs outputs have also been discussed in the context of diffusion-based generation pipelines. \cite{skorokhodov2025improvingdiffusabilityautoencoders} notes that high-frequency components in VAE latent spaces can propagate undesirable artifacts into the final RGB images. In this work, we systematize common and problematic artifacts in VAE reconstructions and propose simple, effective mitigation strategies that require neither extensive changes to the training process nor fundamental modifications of the VAE architecture.

\section{Preliminaries}

\subsection{VAE principles and objectives}\label{sec:vae_theory}

The Variational Autoencoder (VAE) \cite{Kingma2013AutoEncodingVB} is a generative model that learns latent representations of observed data $\mathbf{x}$ through an encoder-decoder architecture. The encoder network parameterizes the approximate posterior distribution $q_\phi(\mathbf{z}|\mathbf{x})$ with weights $\phi$, mapping input data to a latent space typically assumed Gaussian, while the decoder network parameterizes the likelihood $p_\theta(\mathbf{x}|\mathbf{z})$ with weights $\theta$, reconstructing data from latent samples $\mathbf{z}$. The prior over the latent space $p(\mathbf{z})$ is usually chosen as a standard normal distribution $\mathcal{N}(\mathbf{0}, \mathbf{I})$. The VAE optimizes the evidence lower bound (ELBO), which balances reconstruction accuracy and latent space regularization:
\begin{equation}
    \log p(\mathbf{x}) \geq \mathbb{E}_{q_\phi(\mathbf{z}|\mathbf{x})} \log p_\theta(\mathbf{x}|\mathbf{z}) - D_{\text{KL}}\left(q_\phi(\mathbf{z}|\mathbf{x}) \parallel p(\mathbf{z})\right)
\end{equation}
Here, $\log p_\theta(\mathbf{x}|\mathbf{z})$ is the reconstruction term, ensuring that decoded samples match the input data, while the KL divergence $D_{\text{KL}}$ encourages the learned posterior $q_\phi(\mathbf{z}|\mathbf{x})$ to align with the prior $p(\mathbf{z})$.

The VAE training objective combines several loss components that ensure proper latent space regularization and high-quality reconstructions. Modern VAE training setup requires several losses:

\paragraph{KL Divergence Loss \cite{Kingma2013AutoEncodingVB}.} The KL divergence term regularizes the latent space by minimizing the distance between the approximate posterior $q_\phi(\mathbf{z}|\mathbf{x})$ and the prior $p(\mathbf{z}) = \mathcal{N}(\mathbf{0}, \mathbf{I})$:
\begin{equation}
    \mathcal{L}_{\text{KL}} = D_{\text{KL}}\big(q_\phi(\mathbf{z}|\mathbf{x}) \parallel p(\mathbf{z})\big).
\end{equation}
For Gaussian distributions, this has closed-form solution:
\begin{equation}
    \mathcal{L}_{\text{KL}} = -\frac{1}{2}\sum_{j=1}^J \big(1 + \log\sigma_j^2 - \mu_j^2 - \sigma_j^2\big),
\end{equation}
where $\mu_j$ and $\sigma_j$ are the mean and variance of the encoder's posterior approximation.

\paragraph{Reconstruction Loss \cite{Kingma2013AutoEncodingVB}.} The reconstruction term ensures fidelity between input $\mathbf{x}$ and output $\hat{\mathbf{x}} = p_\theta(\mathbf{x}|\mathbf{z})$. For continuous data (MSE):
\begin{equation}
    \mathcal{L}_{\text{recon}} = \mathbb{E}_{q_\phi(\mathbf{z}|\mathbf{x})} \|\mathbf{x} - \hat{\mathbf{x}}\|_2^2.
\end{equation}

\paragraph{Adversarial Loss \cite{Larsen2015AutoencodingBP}.} Regular reconstruction losses in practice often lead to blurry reconstructions. This fact is worsen by the KL term which regularizes the latent space and negatively effects the reconstruction quality. When combined with GAN framework, the adversarial loss improves sample quality:
\begin{equation}
    \mathcal{L}_{\text{adv}} = \mathbb{E}_{q_\phi(\mathbf{z}|\mathbf{x})} \log(1 - D(\hat{\mathbf{x}})),
\end{equation}
where $D$ is a discriminator network trained to distinguish real from generated samples. The discriminator helps automatically detect specific details that VAE fails to reconstruct, unlike the reconstruction loss. Moreover models finetuned with an adversarial loss have proved to have higher alignment with human preferences.

\paragraph{Perceptual Loss \cite{Johnson2016PerceptualLF}.} This is a feature matching loss, which is calculated via a distance metric using deep features $\Phi_l$ from pre-trained networks:
\begin{equation}
    \mathcal{L}_{\text{perc}} = \mathbb{E}_{q_\phi(\mathbf{z}|\mathbf{x})} \|\Phi_l(\mathbf{x}) - \Phi_l(\hat{\mathbf{x}})\|_1.
\end{equation}
A popular choice for the perceptual loss is LPIPS \cite{Zhang2018TheUE}.

\paragraph{Complete Objective.} The total loss combines these components with weighting factors:
\begin{equation}
\mathcal{L}_{\text{total}} = \lambda_{\text{KL}}\mathcal{L}_{\text{KL}} + \lambda_{\text{recon}}\mathcal{L}_{\text{recon}} + \lambda_{\text{adv}}\mathcal{L}_{\text{adv}} + \lambda_{\text{perc}}\mathcal{L}_{\text{perc}}.
\end{equation}

\subsection{VAE architecture}\label{sec:vae_arch}

\begin{figure*}[!t]
  \centering
  \includegraphics[bb=0 0 3454 2163, width=\textwidth]{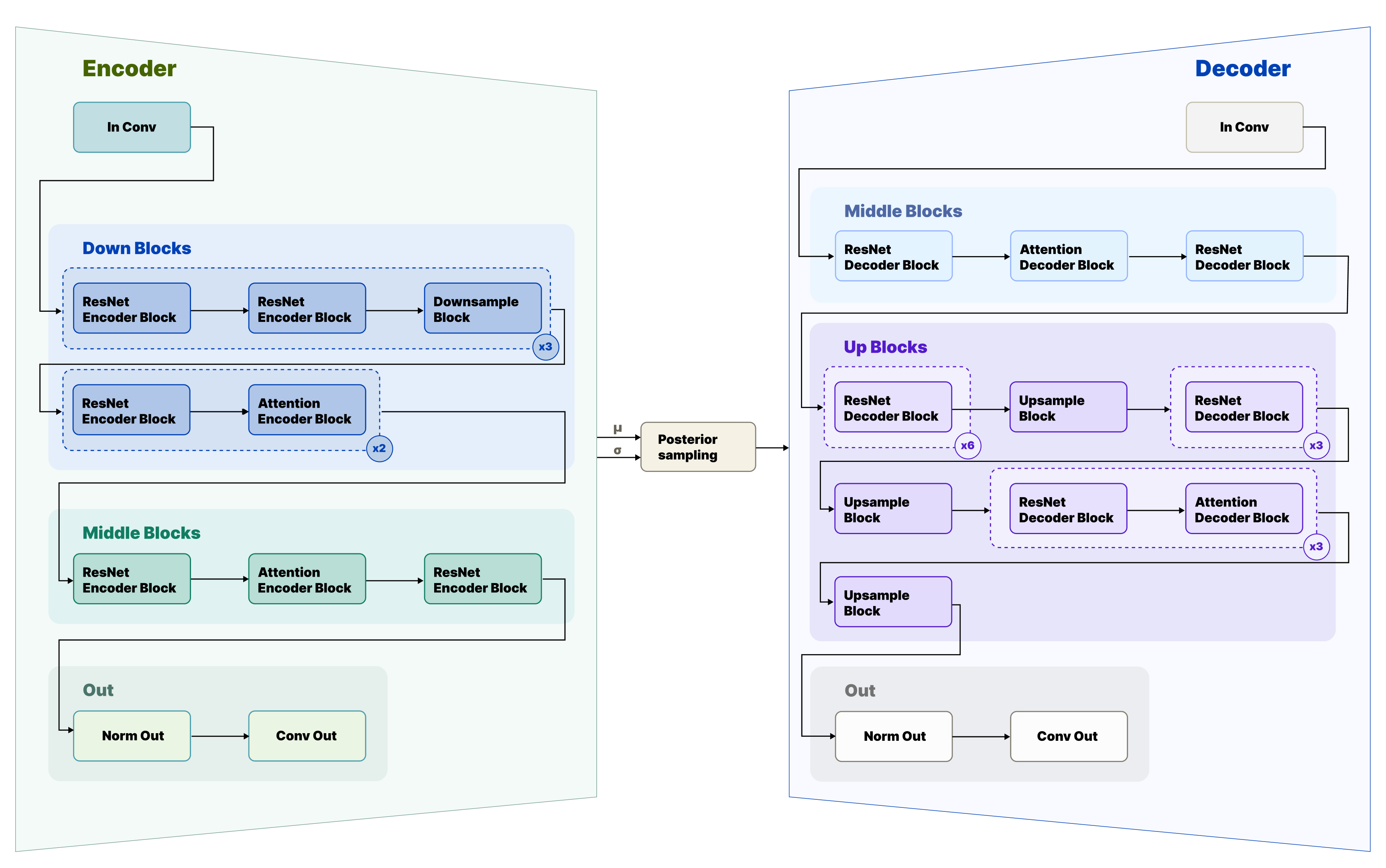}
  \caption{VAE architecture.}
  \label{fig:vae_overview}
\end{figure*}

Our base VAE architecture builds upon the open-source~\footnote{\href{https://github.com/ai-forever/MoVQGAN}{https://github.com/ai-forever/MoVQGAN}} SBER-MoVQGAN model \cite{arkhipkin2024kandinsky30technicalreport}, which itself is derived from VQGAN \cite{esser2020taming} and incorporates spatially conditional normalization from MoVQ \cite{zheng2022movq}. In our implementation, we replace the quantization bottleneck with mean $\mu$ and log-variance $\log \sigma^2$ prediction layers to form a KL-VAE continuous latent space.

Figure \ref{fig:vae_overview} provides a detailed scheme of the KL-VAE structures. Following the design in \cite{Rombach2021HighResolutionIS}, each encoder block consists of a ResNet Block \cite{He2015DeepRL} followed by a downscale module, with some layers additionally incorporating Self-Attention Blocks \cite{Vaswani2017AttentionIA}. The decoder similarly employs ResNet blocks but replaces downscale modules with upscale operations, retaining self-attention where applicable. Detailed ResNet and Self-Attention block schemes are presented in Appendix \ref{appendix:blocks}.

We also introduce minor architectural modifications to address the artifacts described in Section \ref{sec:challenges}. We describe these changes in Section \ref{sec:solutions}, other implementation details can be found in Appendix \ref{appendix:vae}.

\subsection{Latent diffusion VAE}\label{sec:ldm}

Latent diffusion models (LDMs)~\cite{Rombach2021HighResolutionIS} perform diffusion processes in a compressed latent space rather than operating directly on high-dimensional input data (e.g., pixel space for images). By employing a variational autoencoder (VAE) to encode data into a lower-dimensional representation, LDMs achieve significant computational efficiency while maintaining high generation quality.

The VAE encoder transforms an input image $\mathbf{x} \in \mathbb{R}^{H \times W \times C}$ (with height $H$, width $W$, and channels $C$) into a latent representation $\mathbf{z} \in \mathbb{R}^{H/f \times W/f \times c}$, where $f$ is the spatial downscaling factor and $c$ denotes the increased channel dimension for information preservation. In transformer-based implementations~\cite{Peebles2022ScalableDM}, the quadratic complexity of full self-attention with respect to latents size makes spatial dimension operations particularly costly. Channel dimension adjustments, however, can be efficiently implemented through linear projections with minimal computational overhead.

In this work, we further investigate the suitability of our enhanced image reconstruction VAE in a latent diffusion pipeline for text-to-image generation. We employ a diffusion transformer based on LI-DiT \cite{ma2024exploring}, training it with our frozen VAE, and present the results in Section \ref{sec:generation_results}.

\begin{figure*}[t!]
    \centering
    \begin{subfigure}[t]{0.32\textwidth}
        \centering
        \includegraphics[bb=0 0 1028 1028, width=\textwidth]{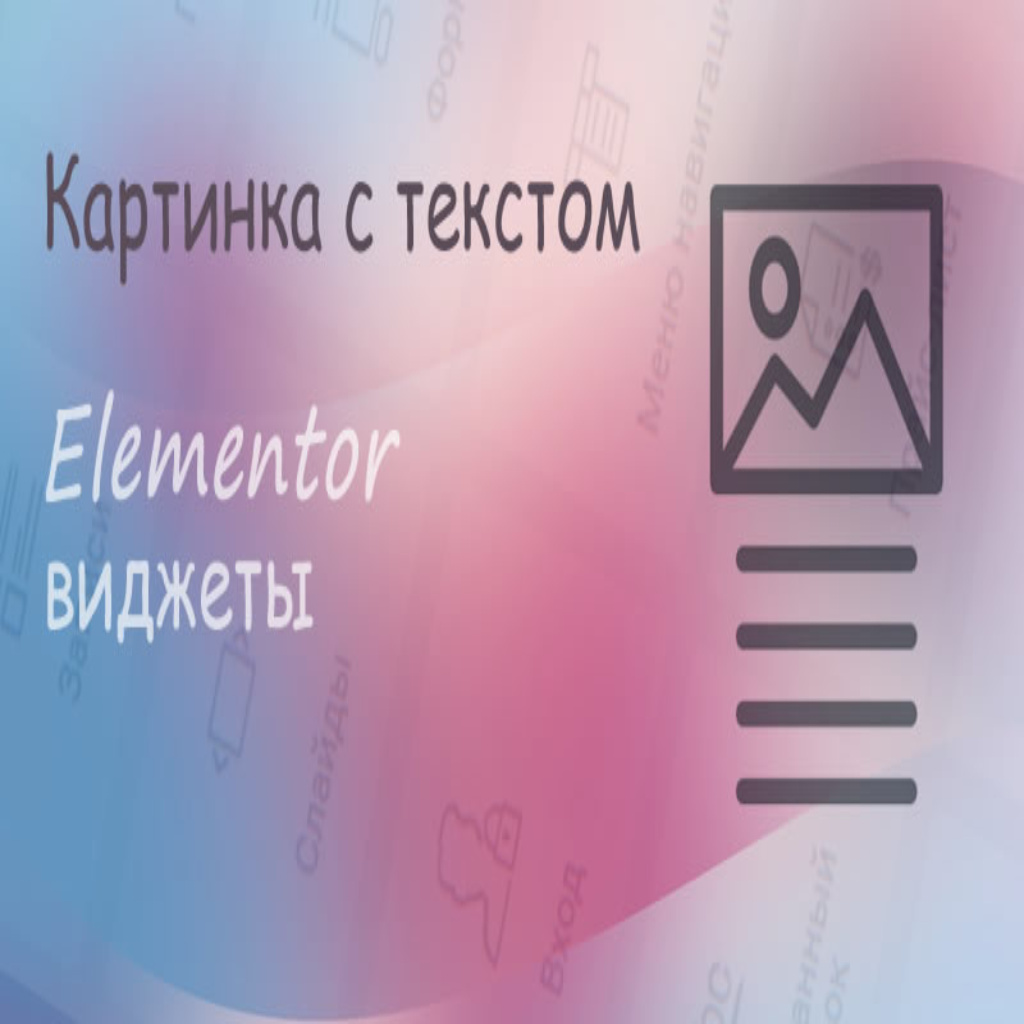}
        \caption{Original image}
        \label{pic:art1}
    \end{subfigure}%
    ~ 
    \begin{subfigure}[t]{0.32\textwidth}
        \centering
        \includegraphics[bb=0 0 220 220, width=\textwidth]{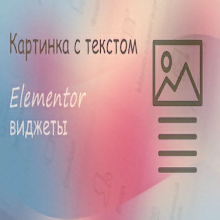}
        \caption{Color shift: noticeable color change throughout the picture}
        \label{pic:art2}
    \end{subfigure}
     ~ 
    \begin{subfigure}[t]{0.32\textwidth}
        \centering
        \includegraphics[bb=0 0 540 540, width=\textwidth]{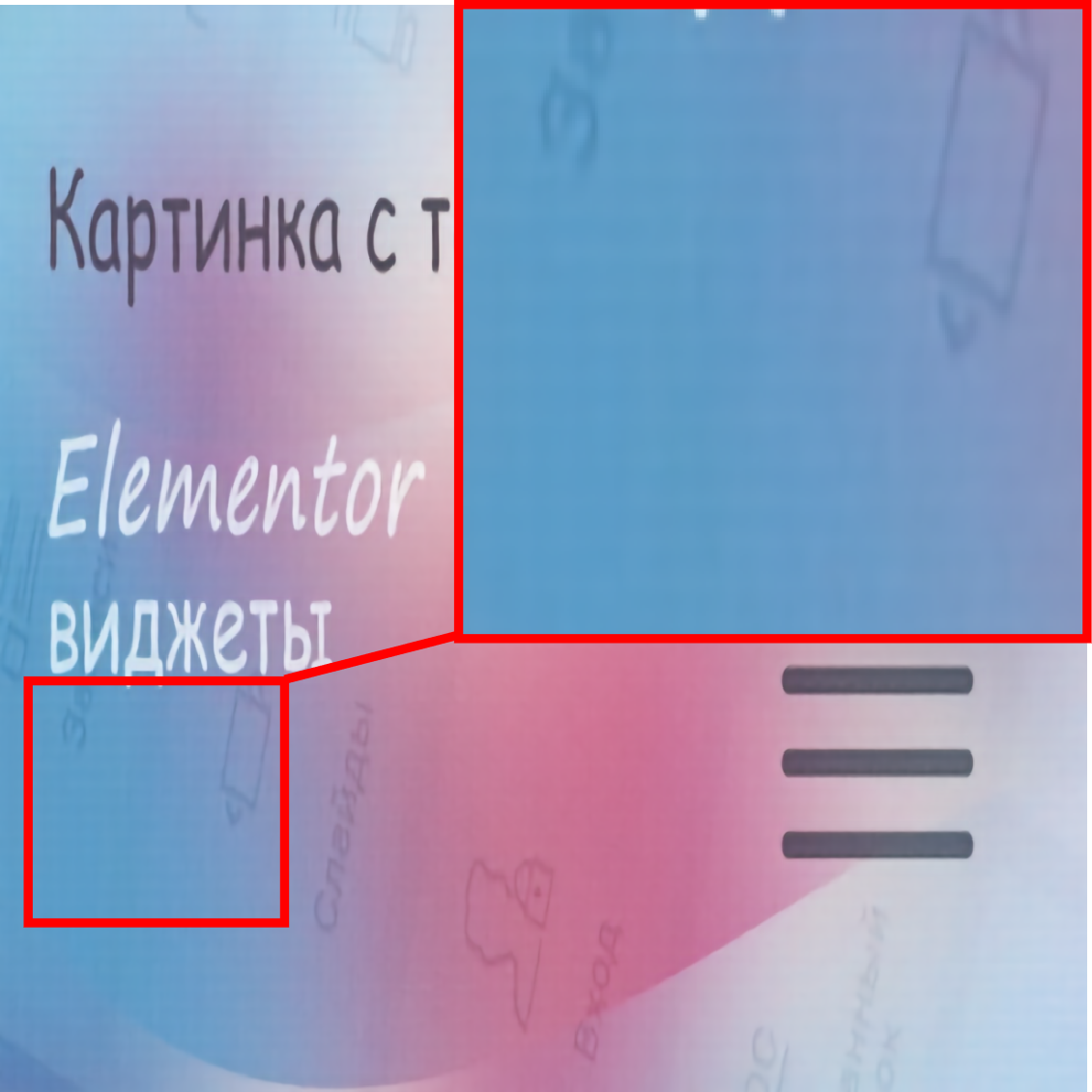}
        \caption{Grid: fine mesh over the entire image}
        \label{pic:art3}
    \end{subfigure}
    ~
    \begin{subfigure}[t]{0.32\textwidth}
        \centering
        \includegraphics[bb=0 0 133 133, width=\textwidth]{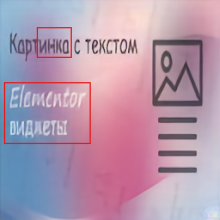}
        \caption{Blur: some elements of pictures are blurry}
        \label{pic:art4}
    \end{subfigure}%
    ~ 
    \begin{subfigure}[t]{0.32\textwidth}
        \centering
        \includegraphics[bb=0 0 133 133, width=\textwidth]{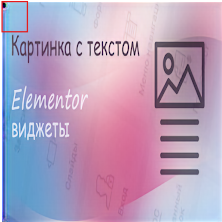}
        \caption{Corner artifact: bright unnatural color change in the corner and also along the edge of the picture}
        \label{pic:art5}
    \end{subfigure}
     ~ 
    \begin{subfigure}[t]{0.32\textwidth}
        \centering
        \includegraphics[bb=0 0 308 308, width=\textwidth]{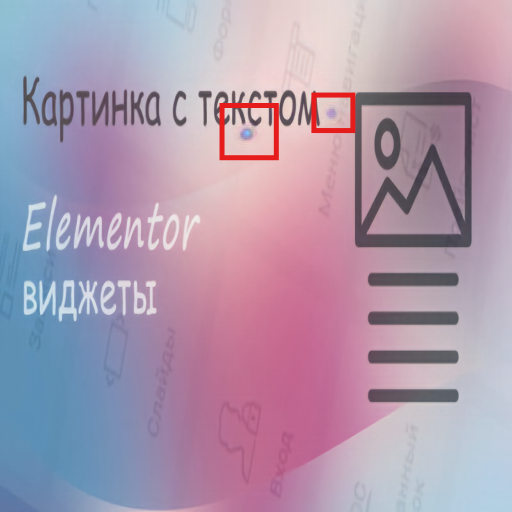}
        \caption{Droplet artifact: bright unnatural color change in the random place on the picture}
        \label{pic:art6}
    \end{subfigure}
    \caption{Reconstruction artifacts resulting from the unimproved VAE model.}\label{pic:artifacts}
\end{figure*}

\section{Image VAE challenges}\label{sec:challenges}

Researchers frequently encounter challenges when training VAEs, yet most studies only report final configurations while omitting the methodological details of how they were achieved. Firstly, the difference in data collection can cause irreproducible results even for the full repetition of the training pipeline. Secondly, the extensive number of  hyperparameters inherent in VAE training makes the learning process highly complex. Researchers must determine an appropriate combination of loss functions and carefully select hyperparameters for each training setup, for example, when increasing image resolution. Due to the interdependence of hyperparameters, suboptimal tuning can result in significant training difficulties and artifacts in reconstructed images. Finally, incorporating a discriminator in VAE training can introduce additional challenges, such as instability, vanishing gradients, and mode collapse. All aforementioned issues often manifest as specific defects in VAE reconstructions. We argue that these defects can be systematically categorized based on their visual characteristics and underlying causes. Below, we provide an overview of common artifacts and their potential origins.

\begin{figure*}[b!]
    \centering
    \begin{subfigure}[t]{0.215\textwidth}
        \centering
        \includegraphics[bb=0 0 300 300, width=\textwidth]{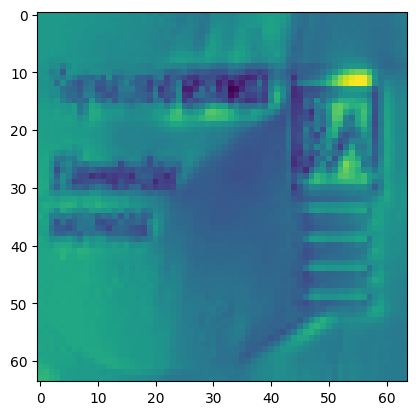}
        \caption{Output from the first Decoder Self-Attention Block}
    \end{subfigure}%
    ~ 
    \begin{subfigure}[t]{0.22\textwidth}
        \centering
        \includegraphics[bb=0 0 307 307, width=\textwidth]{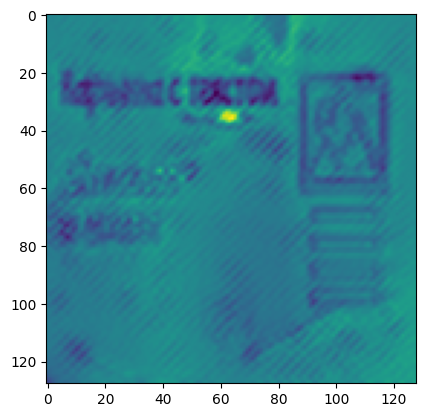}
        \caption{After the ResNet-Blocks, before second Upsample Block}
    \end{subfigure}
     ~ 
    \begin{subfigure}[t]{0.22\textwidth}
        \centering
        \includegraphics[bb=0 0 307 307, width=\textwidth]{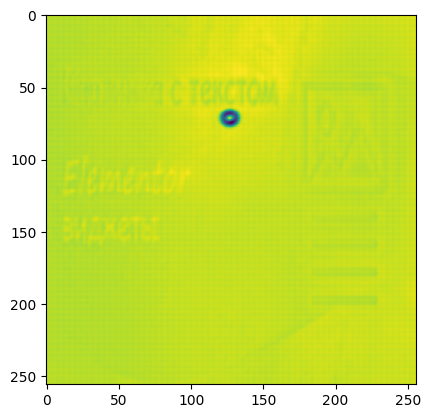}
        \caption{Before third Upsample Block}
    \end{subfigure}
    ~
    \begin{subfigure}[t]{0.22\textwidth}
        \centering
        \includegraphics[bb=0 0 307 307, width=\textwidth]{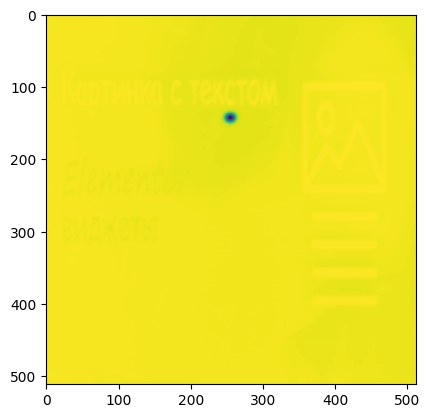}
        \caption{Norms of output}
    \end{subfigure}
    ~
    \begin{subfigure}[t]{0.035\textwidth}
        \centering
        \includegraphics[bb=0 0 60 300, width=\textwidth]{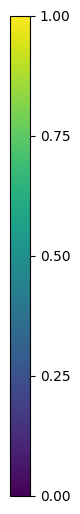}
    \end{subfigure}
    \caption{Droplet artifact formation. Activation norms grow on several consecutive Decoder layers. }\label{pic:droplet_activations}
\end{figure*}

\paragraph{Color shift.}
The first artifact while training VAE with adversarial loss is a color shift. It manifests itself in the fact that the color of all generated images is moved towards one color. In our case all the pictures had a yellow tint. The example of this problem is presented in Figure~\ref{pic:art2}. This artifact appear at the early stages of training. One of the possible reasons of it is the adversarial training setup. Such problem can be considered as a kind of mode collapse.

\paragraph{Grid.}
One more unpleasant attribute which can occur in the reconstructed images while training VAE with the discriminator. The example of this problem is presented in Figure~\ref{pic:art3}. In our experiments, this problem arose when the adversarial loss had too much weight.

\paragraph{Blur.}
The most common artifact while training the VAE. Unlike the previous problems, its appearance is associated with the regularization of the latent space during training of the VAE with the KL divergence loss. This part of loss pushes posterior distribution to the $\mathcal{N}(\textbf{0}, \textbf{I})$ so the hight weight for it can create too simple distribution of the latent variables and blurred reconstructions. The example of this problem is presented in Figure~\ref{pic:art4}. 

\paragraph{Corner artifact.}
The next two problems are connected with the model architecture design. The first of them is the corner artifact. This artifact looks like a frame of blur around the edges of the image and presented in the Figure~\ref{pic:art5}. Explanation for this can be found in the convolutional layers in the architecture and particularly in padding used in them. In our experiments we found out that zero padding cause this problem.

\paragraph{Droplet artifact.}
This peculiarity was also mentioned in the~\cite{polyak2024moviegen}. The authors pointed out that this problem caused by the high norms in latent codes in certain spatial locations. After decoding these high norms produce the bright spots in the pixel space. In our work we also faced with this problem, the example is presented at the Figure~\ref{pic:art6}. Also the Figure~\ref{pic:droplet_activations} shows formation process of this artifact by displaying the activations norms for the different layers of the VAE decoder. To mitigate this problem the authors of the~\cite{polyak2024moviegen} invented an additional outlier loss, which penalizes the model for latent values far from the mean. We propose to solve this problem in other, simpler way.

\section{Methods}\label{sec:methods}

This section describes the data selection and preprocessing methods, and presents our findings about on mitigating the problems of VAE training depicted in Section \ref{sec:challenges}. 

\subsection{Data preprocessing}\label{sec:preprocessing}

Data quality is a critical factor influencing the performance of VAE. Specifically, the training dataset should consist of high-resolution images exhibiting a substantial amount of high-frequency detail.

Another key consideration is data preprocessing, which typically involves resizing and cropping images to reduce their resolution. However, our experiments demonstrate that directly cropping high-resolution images (e.g., 720p or HD) to a lower resolution (e.g., 240p) can adversely affect VAE training. Such transformations yield images that are effectively magnified regions of the originals, lacking sufficient high-frequency information. Consequently, these images do not possess the representational complexity necessary for training a high-quality VAE. Instead, a more effective approach involves first proportionally resizing a large image to an intermediate resolution (e.g., $\sim$480p) before cropping to the target resolution (e.g., 240p).

The resizing method also significantly impacts performance. Our experiments indicate that the default nearest-neighbor method yields suboptimal results, whereas the bicubic method produces superior outcomes. These findings align with established results from prior work~\cite{Han2013, Bhatt}. Notably, training a VAE on lower-resolution images (e.g., 240p) can remain effective for operation at higher resolutions (e.g., 480p or 720p), as the convolutional receptive field remains unchanged. Thus, the VAE learns to compress the image by a factor of $n$ along each side in the same manner as with smaller resolutions.

Our complete training dataset consists of 100 million images sourced from the publicly available LAION Aesthetic HighRes dataset, a subset of LAION-5B~\cite{schuhmann2022laion}. All images in this dataset have HD resolution and an associated aesthetic score. We selected only images with a score $> 4.5$, resized them from HD to 480p, and then cropped them to 240p.

\subsection{Challenges solutions}\label{sec:solutions}

\paragraph{Color shift.}
This problem is very common and can occur in many different situations. Often this artifact reveals itself in the early steps of training and indicates that the model has not been fully trained. This problems is solved in two ways: either prolonged training of the whole model or only training the decoder part.

\paragraph{Grid.}
As it was said in previous section this unpleasant artifact that occurred due to the large weight on the discriminator. So, this problem can be reduced by the lowering this weight. 

\paragraph{Blur.}
As this the most common problem with VAE, arising due to the very essence of KL regularization of the latent space. It can be solved by reducing the weight for the regularization term and increasing the weight on the discriminator. But such solution contradict with the previous artifact. Therefore it is very important to choose the right weight for the discriminator and the KL regularization term. In our case, decreasing weight of KL divergence by 10 times solved the problem.

\paragraph{Corner artifact.}
This undesirable artifact as it was said above appears due to the zero padding in the convolutions layers of the VAE. Actually, one can not to face with such issue  because many open source architectures use this type of padding and do not obtain similar problem. But in our work we uncover that this picture deterioration is completely cured by changing the padding type to reflect.

\paragraph{Droplet artifact.}
At first this artifact appeared in the corner and we assumed that it also appeared due to incorrect padding and that we would get rid of it with the previous artifact after padding changing. But after that the dot began to move throughout the whole picture. As it was already mentioned the real reason is the high norms in activations during decoding. Unlike the~\cite{polyak2024moviegen} we decided not to add the new loss term, as it would make the total loss more complicated and hard to adjust its weights, but make changes to the architecture Of the model. We use the method proposed in~\cite{zheng2022movq} named Spatially Conditional Normalization. The authors used this normalization to modulate the quantized vectors so as to insert spatially variant information to the embedded index maps. In this way, they achieved a reduction in the number of artifacts that arise due to encoding similar patches into one latent codebook vector. For us this layer has two main benefits. First, it acts as a supplementary normalization layer. Second, it provides additional spatial interactions between parts of the image. And although we are not working with a discrete latent space, using this layer helps us get rid of the droplet artifact.  

\subsection{Decoder finetuning}
Another sufficient step to improve VAE reconstruction quality is a freezing of encoder. This training stage can further improve quality of trained model and help to get rid of other artifacts. In our experiments, an additional training phase with a fixed encoder helped us further increase the FID metric at a late phase of training. Similar strategy was used for example in~\cite{audiovae} where the stage with frozen encoder was even longer then full VAE training stage.

\begin{figure*}[t]
  \centering
  \includegraphics[bb=0 0 1726 795, width=\textwidth]{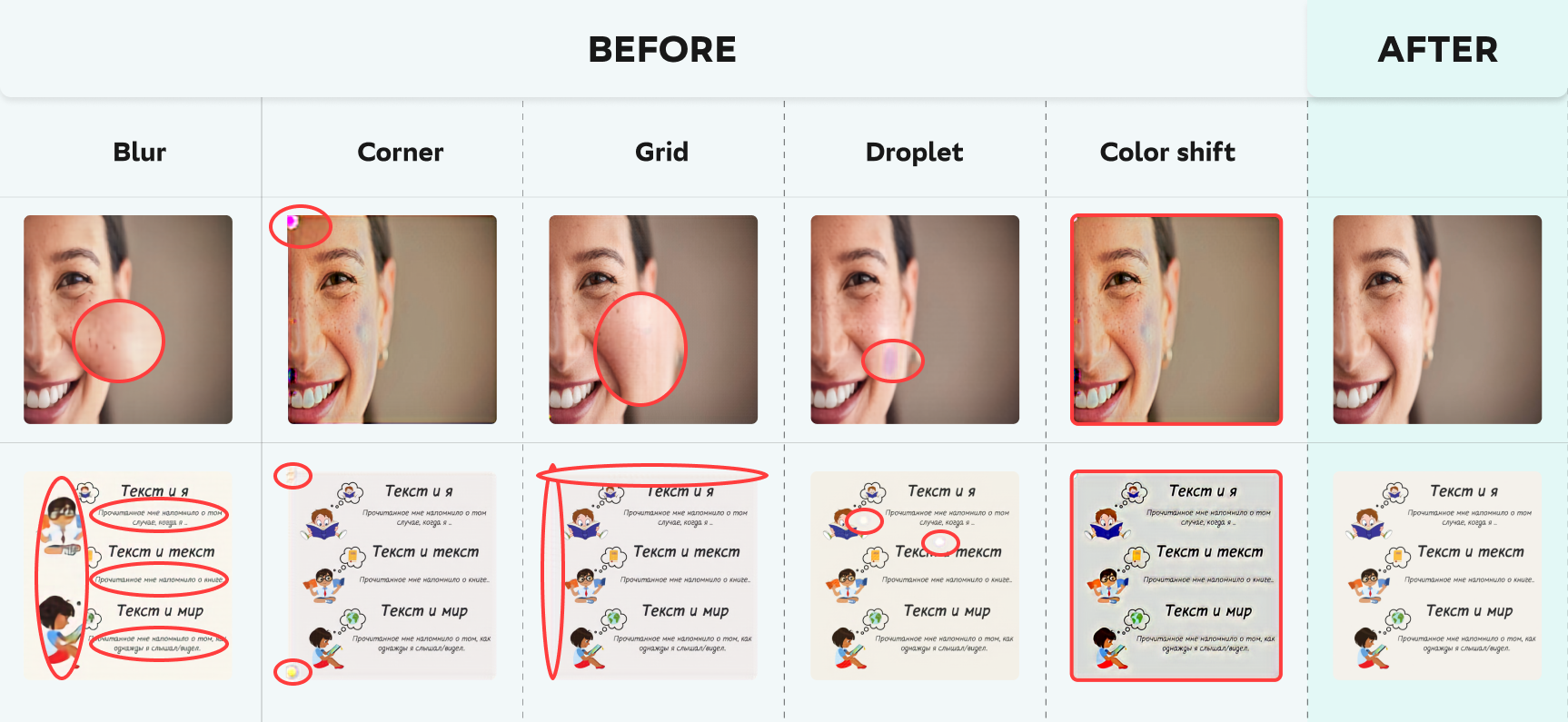}
  \caption{The results of our methods for addressing VAE reconstruction artifacts. Our proposed approach effectively eliminates many problems, ensuring high-quality reconstructions.}
  \label{fig:artifacts_mitigation}
\end{figure*}

\begin{figure*}[t]
  \centering
  \includegraphics[bb=0 0 2300 2038, width=\textwidth]{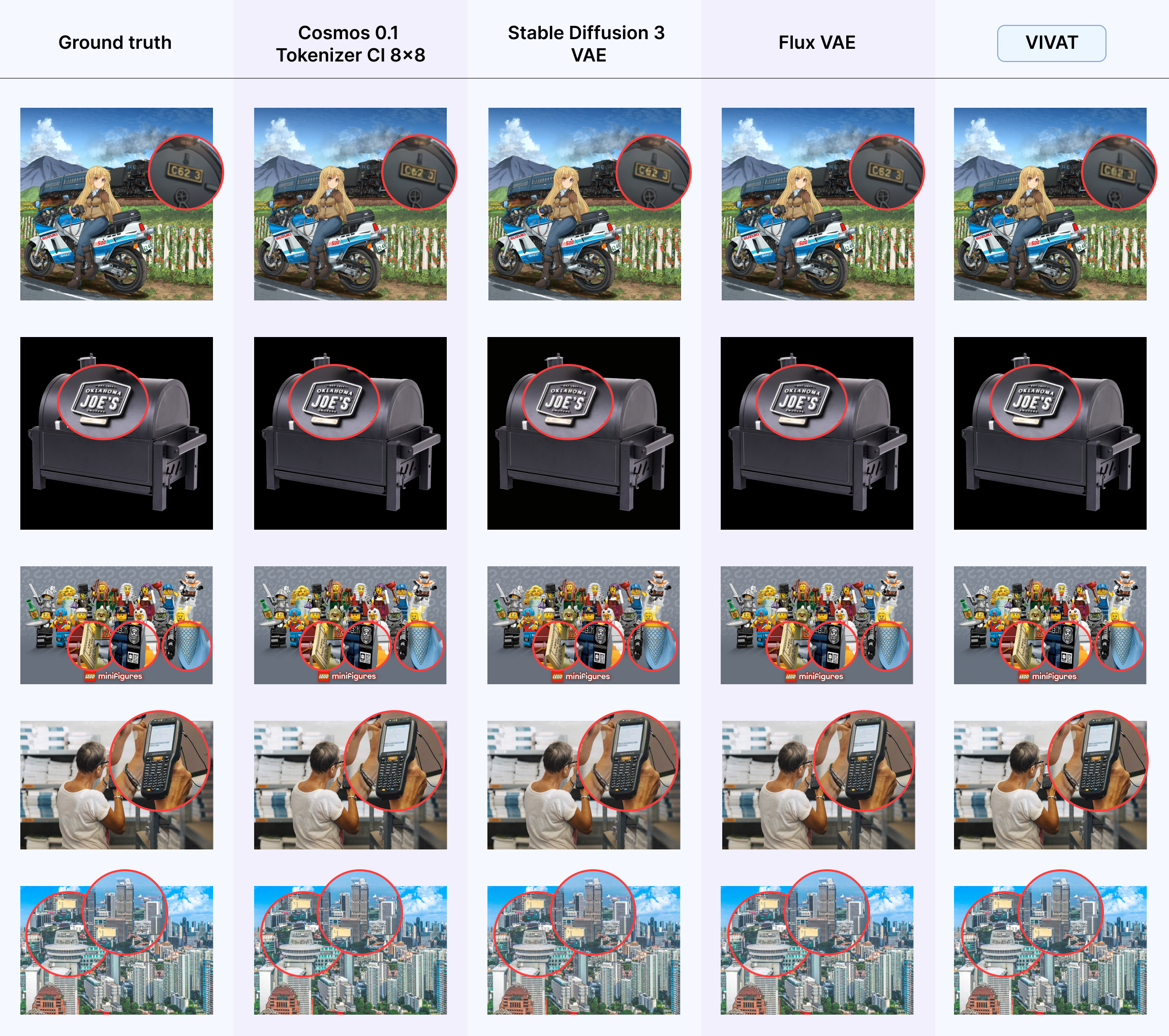}
  \caption{The results of image reconstruction using different models. A zoom-in shows that our approach, based on simple heuristics, leads to the results compared to state-of-the-art models. By artifacts mitigation, it is possible to achieve higher quality in the reconstruction of small details and text.}
  \label{fig:comparison}
\end{figure*}

\section{Results}\label{sec:results}

\subsection{Artifact mitigation}\label{sec:artifact_mitigation}

We applied the techniques described in Section \ref{sec:solutions} and trained VAE model on the dataset described in Section \ref{sec:preprocessing} and with the training parameters that can be found in Appendix \ref{appendix:vae}. We conducted a qualitative comparison between new model and the base VAE model described in Section \ref{sec:vae_arch}. The results are presented in Figure \ref{fig:artifacts_mitigation}. As can be clearly seen, our approach effectively eliminates all artifacts listed in Section \ref{sec:challenges}. More examples can be found in Appendix \ref{appendix:results}.
 
\subsection{Image reconstruction}\label{sec:reconstruction_results}

To assess the reconstruction performance of our approach, we employ two standard metrics: peak signal-to-noise ratio (PSNR) and structural similarity index measure (SSIM). We evaluate these metrics on multiple benchmark datasets, including ImageNet\footnote{\href{https://www.image-net.org/challenges/LSVRC/2012/}{https://www.image-net.org/challenges/LSVRC/2012/}} \cite{5206848} at resolutions of $256 \times 256$ and $512 \times 512$, MS COCO 2017 Test\footnote{\href{https://cocodataset.org/}{https://cocodataset.org/}} \cite{lin2015microsoftcococommonobjects} at $512 \times 512$, and FFHQ\footnote{\href{https://github.com/NVlabs/ffhq-dataset}{https://github.com/NVlabs/ffhq-dataset}} \cite{karras2019stylebasedgeneratorarchitecturegenerative} at $1024 \times 1024$.

For comparison, we test our model against several state-of-the-art autoencoders: DC-VAE \cite{chen2024deep}, 
Cosmos-0.1-Tokenizer-CI8$\times$8\footnote{\href{https://huggingface.co/nvidia/Cosmos-0.1-Tokenizer-CI8x8}{https://huggingface.co/nvidia/Cosmos-0.1-Tokenizer-CI8x8}} \cite{nvidia2025cosmosworldfoundationmodel}, Flux VAE\footnote{\href{https://huggingface.co/black-forest-labs/FLUX.1-dev}{https://huggingface.co/black-forest-labs/FLUX.1-dev}} \cite{flux2024}, and the VAE from Stable Diffusion 3 \cite{esser2024scalingrectifiedflowtransformers}. The quantitative results, presented in Table~\ref{tab:recon_metrics}, demonstrate that our method achieves state-of-the-art performance in image reconstruction, attaining the highest -- or competitively equivalent -- scores in both SSIM and PSNR across all evaluated datasets. This quantitative superiority is further supported by qualitative evaluation, as illustrated in Figure \ref{fig:comparison}, where our method exhibits competitive results compared to the best approaches.

\begin{table*}[ht]
\centering
\begin{tabular}{|l|c|c|c|c|}
\hline
Model & ImageNet & ImageNet & MS COCO  & FFHQ\\
      & $256 \times 256$ & $512 \times 512$ & 2017 Test & $1024 \times 1024$\\
      &         &         & $512 \times 512$ & \\
\hline
& \multicolumn{4}{|c|}{PSNR$\uparrow$} \\
\hline
DC-VAE & 23.7 & 26.15 & 25.38 & 31.34 \\
Cosmos-0.1-Tokenizer-CI8$\times$8 & 30.54 & 33.76 & 32.28 & \underline{39.12}\\
Flux VAE & \underline{31.09} & \textbf{33.99} & \textbf{32.65} & 38.14\\
Stable Diffusion 3 VAE & 29.58 & 32.07 & 30.94 & 36.16\\
VIVAT (Our) & \textbf{31.25} & \underline{33.82} & \underline{32.61} & \textbf{39.35}\\
\hline

& \multicolumn{4}{|c|}{SSIM$\uparrow$} \\
\hline
DC-VAE  &  0.65 & 0.71 & 0.70 & 0.83\\
Cosmos-0.1-Tokenizer-CI8x8 & 0.88 & 0.91 & 0.90 & 0.95\\
Flux VAE & \underline{0.89} & \textbf{0.93} & \textbf{0.91} & \textbf{0.96}\\
Stable Diffusion 3 VAE & 0.86 & 0.89 &  0.87 & 0.93\\
VIVAT (Our) & \textbf{0.90} & \underline{0.92} & \textbf{0.91} & \textbf{0.96}\\
\hline
\end{tabular}
\caption{
Image reconstruction results.
The best values are \textbf{highlighted}, and the second-best values are \underline{underlined}.}
\label{tab:recon_metrics}
\end{table*}

\begin{figure*}[ht]
  \centering
  \includegraphics[bb=0 0 496 208, width=0.8\textwidth]{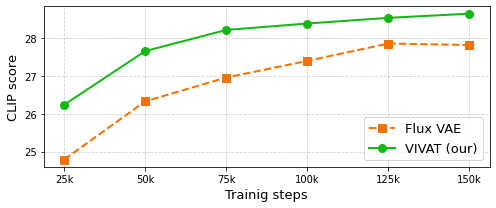}
  \caption{The CLIP score on the validation dataset during the training process of the diffusion transformer with image embeddings obtained from Flux VAE or our method VIVAT. Our method shows a stable improvement in text-image alignment and outperforms the competitive method.}
  \label{fig:clip_score}
\end{figure*}

\subsection{Text-to-image generation}\label{sec:generation_results}

To assess the integration of our VAE with latent diffusion pipeline, we train a Diffusion Transformer (DiT), which is similar to LI-DiT model \cite{ma2024exploring}. The architecture consists of two principal components:

\begin{itemize}
    \item Token Refiner that processes textual embeddings through self-attention and feed-forward layers with Adaptive Layer Normalization (AdaNorm) \cite{xu2019understandingimprovinglayernormalization} for temporal conditioning;
    \item Cross-Attention Block that handles visual embeddings via self-attention while enabling text conditioning through cross-attention mechanisms and employing AdaNorm for temporal information. 
\end{itemize}

For comparative evaluation between VIVAT and Flux VAE \cite{flux2024}, we trained a 2 billion parameter DiT model for 130 thousand steps on $256 \times 256$ resolution images with a batch size of 1024. Generation quality was quantified using the CLIP-score metric \cite{Hessel2021CLIPScoreAR} to measure text-image alignment. Figure \ref{fig:clip_score} shows that our method allows for a stable improvement in quality during the training process and surpasses the pipeline with the same diffusion transformer, but using Flux VAE. Details of the architecture and training of the diffusion transformer can be found in Appendix \ref{appendix:dit}.

\section{Conclusion}

In this work, we identified and addressed key challenges in KL-VAE training through a systematic analysis of artifacts and their mitigation. By refining loss weightings, architectural details, and training strategies, VIVAT achieves superior reconstruction quality and enhances performance of the latent diffusion pipeline for text-to-image generation. Our approach demonstrates that significant improvements can be made without departing from the classical VAE framework, offering a practical and scalable solution for real-world applications. The success of VIVAT underscores the untapped potential of traditional autoencoding methods when combined with targeted optimizations. Future work could explore the extension of these principles to other generative tasks and architectures, including VAE for video generation pipelines.

\section*{Acknowledgments}

The authors express the gratitude to Evelina Sidorova for her assistance in preparing the illustrations for this paper.

\bibliographystyle{ACM-Reference-Format}
\bibliography{references}

@String{Computer = "{IEEE} Computer" }

@String{Springer = "Springer-Verlag" }

@misc{polyak2024moviegen,
      title={Movie Gen: A Cast of Media Foundation Models}, 
      author={Adam Polyak and others},
      year={2024},
      eprint={2410.13720},
      archivePrefix={arXiv},
      primaryClass={cs.CV},
      url={https://arxiv.org/abs/2410.13720}, 
}

@misc{Rombach2021HighResolutionIS,
      title={High-Resolution Image Synthesis with Latent Diffusion Models}, 
      author={Robin Rombach and Andreas Blattmann and Dominik Lorenz and Patrick Esser and Björn Ommer},
      year={2022},
      eprint={2112.10752},
      archivePrefix={arXiv},
      primaryClass={cs.CV},
      url={https://arxiv.org/abs/2112.10752}, 
}

@misc{Zhang2018TheUE,
      title={The Unreasonable Effectiveness of Deep Features as a Perceptual Metric}, 
      author={Richard Zhang and Phillip Isola and Alexei A. Efros and Eli Shechtman and Oliver Wang},
      year={2018},
      eprint={1801.03924},
      archivePrefix={arXiv},
      primaryClass={cs.CV},
      url={https://arxiv.org/abs/1801.03924}, 
}

@misc{Larsen2015AutoencodingBP,
      title={Autoencoding beyond pixels using a learned similarity metric}, 
      author={Anders Boesen Lindbo Larsen and Søren Kaae Sønderby and Hugo Larochelle and Ole Winther},
      year={2016},
      eprint={1512.09300},
      archivePrefix={arXiv},
      primaryClass={cs.LG},
      url={https://arxiv.org/abs/1512.09300}, 
}

@misc{Peebles2022ScalableDM,
      title={Scalable Diffusion Models with Transformers}, 
      author={William Peebles and Saining Xie},
      year={2023},
      eprint={2212.09748},
      archivePrefix={arXiv},
      primaryClass={cs.CV},
      url={https://arxiv.org/abs/2212.09748}, 
}

@inproceedings{zheng2022movq,
    author = {Zheng, Chuanxia and Vuong, Long Tung and Cai, Jianfei and Phung, Dinh},
    title = {MoVQ: modulating quantized vectors for high-fidelity image generation},
    year = {2022},
    isbn = {9781713871088},
    publisher = {Curran Associates Inc.},
    address = {Red Hook, NY, USA},
    booktitle = {Proceedings of the 36th International Conference on Neural Information Processing Systems},
    articleno = {1701},
    numpages = {14},
    location = {New Orleans, LA, USA},
    series = {NIPS '22}
}

@inproceedings{Vaswani2017AttentionIA,
author = {Vaswani, Ashish and Shazeer, Noam and Parmar, Niki and Uszkoreit, Jakob and Jones, Llion and Gomez, Aidan N. and Kaiser, \L{}ukasz and Polosukhin, Illia},
title = {Attention is all you need},
year = {2017},
isbn = {9781510860964},
publisher = {Curran Associates Inc.},
address = {Red Hook, NY, USA},
booktitle = {Proceedings of the 31st International Conference on Neural Information Processing Systems},
pages = {6000–6010},
numpages = {11},
location = {Long Beach, California, USA},
series = {NIPS'17}
}

@misc{He2015DeepRL,
      title={Deep Residual Learning for Image Recognition}, 
      author={Kaiming He and Xiangyu Zhang and Shaoqing Ren and Jian Sun},
      year={2015},
      eprint={1512.03385},
      archivePrefix={arXiv},
      primaryClass={cs.CV},
      url={https://arxiv.org/abs/1512.03385}, 
}

@misc{Johnson2016PerceptualLF,
      title={Perceptual Losses for Real-Time Style Transfer and Super-Resolution}, 
      author={Justin Johnson and Alexandre Alahi and Li Fei-Fei},
      year={2016},
      eprint={1603.08155},
      archivePrefix={arXiv},
      primaryClass={cs.CV},
      url={https://arxiv.org/abs/1603.08155}, 
}

@misc{Kingma2013AutoEncodingVB,
      title={Auto-Encoding Variational Bayes}, 
      author={Diederik P Kingma and Max Welling},
      year={2013},
      eprint={1312.6114},
      archivePrefix={arXiv},
      primaryClass={stat.ML},
      url={https://arxiv.org/abs/1312.6114}, 
}

@InProceedings{pmlr-v32-rezende14,
  title = 	 {Stochastic Backpropagation and Approximate Inference in Deep Generative Models},
  author = 	 {Rezende, Danilo Jimenez and Mohamed, Shakir and Wierstra, Daan},
  booktitle = 	 {Proceedings of the 31st International Conference on Machine Learning},
  pages = 	 {1278--1286},
  year = 	 {2014},
  editor = 	 {Xing, Eric P. and Jebara, Tony},
  volume = 	 {32},
  series = 	 {Proceedings of Machine Learning Research},
  address = 	 {Bejing, China},
  month = 	 {06},
  publisher =    {PMLR},
  url = 	 {https://proceedings.mlr.press/v32/rezende14.html}
}

@InProceedings{10.1007/978-3-319-46493-0_47,
author="Yan, Xinchen
and Yang, Jimei
and Sohn, Kihyuk
and Lee, Honglak",
editor="Leibe, Bastian
and Matas, Jiri
and Sebe, Nicu
and Welling, Max",
title="Attribute2Image: Conditional Image Generation from Visual Attributes",
booktitle="Computer Vision -- ECCV 2016",
year="2016",
publisher="Springer International Publishing",
address="Cham",
pages="776--791",
isbn="978-3-319-46493-0"
}

@misc{10.5555/3454287.3455618,
      title={Generating Diverse High-Fidelity Images with VQ-VAE-2}, 
      author={Ali Razavi and Aaron van den Oord and Oriol Vinyals},
      year={2019},
      eprint={1906.00446},
      archivePrefix={arXiv},
      primaryClass={cs.LG},
      url={https://arxiv.org/abs/1906.00446}, 
}

@misc{vahdat2020NVAE,
      title={NVAE: A Deep Hierarchical Variational Autoencoder}, 
      author={Arash Vahdat and Jan Kautz},
      year={2021},
      eprint={2007.03898},
      archivePrefix={arXiv},
      primaryClass={stat.ML},
      url={https://arxiv.org/abs/2007.03898}, 
}

@misc{esser2020taming,
      title={Taming Transformers for High-Resolution Image Synthesis}, 
      author={Patrick Esser and Robin Rombach and Björn Ommer},
      year={2020},
      eprint={2012.09841},
      archivePrefix={arXiv},
      primaryClass={cs.CV}
}

@misc{yu2022vectorquantizedimagemodelingimproved,
      title={Vector-quantized Image Modeling with Improved VQGAN}, 
      author={Jiahui Yu and Xin Li and Jing Yu Koh and Han Zhang and Ruoming Pang and James Qin and Alexander Ku and Yuanzhong Xu and Jason Baldridge and Yonghui Wu},
      year={2022},
      eprint={2110.04627},
      archivePrefix={arXiv},
      primaryClass={cs.CV},
      url={https://arxiv.org/abs/2110.04627}, 
}

@misc{podell2023sdxlimprovinglatentdiffusion,
      title={SDXL: Improving Latent Diffusion Models for High-Resolution Image Synthesis}, 
      author={Dustin Podell and Zion English and Kyle Lacey and Andreas Blattmann and Tim Dockhorn and Jonas Müller and Joe Penna and Robin Rombach},
      year={2023},
      eprint={2307.01952},
      archivePrefix={arXiv},
      primaryClass={cs.CV},
      url={https://arxiv.org/abs/2307.01952}, 
}

@misc{chen2024deep,
      title={Deep Compression Autoencoder for Efficient High-Resolution Diffusion Models}, 
      author={Junyu Chen and Han Cai and Junsong Chen and Enze Xie and Shang Yang and Haotian Tang and Muyang Li and Yao Lu and Song Han},
      year={2025},
      eprint={2410.10733},
      archivePrefix={arXiv},
      primaryClass={cs.CV},
      url={https://arxiv.org/abs/2410.10733}, 
}

@misc{chen2023pixartalpha,
      title={PixArt-$\alpha$: Fast Training of Diffusion Transformer for Photorealistic Text-to-Image Synthesis}, 
      author={Junsong Chen and Jincheng Yu and Chongjian Ge and Lewei Yao and Enze Xie and Yue Wu and Zhongdao Wang and James Kwok and Ping Luo and Huchuan Lu and Zhenguo Li},
      year={2023},
      eprint={2310.00426},
      archivePrefix={arXiv},
      primaryClass={cs.CV}
}

@misc{esser2024scalingrectifiedflowtransformers,
      title={Scaling Rectified Flow Transformers for High-Resolution Image Synthesis}, 
      author={Patrick Esser and Sumith Kulal and Andreas Blattmann and Rahim Entezari and Jonas Müller and Harry Saini and Yam Levi and Dominik Lorenz and Axel Sauer and Frederic Boesel and Dustin Podell and Tim Dockhorn and Zion English and Kyle Lacey and Alex Goodwin and Yannik Marek and Robin Rombach},
      year={2024},
      eprint={2403.03206},
      archivePrefix={arXiv},
      primaryClass={cs.CV},
      url={https://arxiv.org/abs/2403.03206}, 
}

@misc{arkhipkin2024kandinsky30technicalreport,
      title={Kandinsky 3.0 Technical Report}, 
      author={Vladimir Arkhipkin and Andrei Filatov and Viacheslav Vasilev and Anastasia Maltseva and Said Azizov and Igor Pavlov and Julia Agafonova and Andrey Kuznetsov and Denis Dimitrov},
      year={2024},
      eprint={2312.03511},
      archivePrefix={arXiv},
      primaryClass={cs.CV},
      url={https://arxiv.org/abs/2312.03511}, 
}

@inproceedings{vladimir-etal-2024-kandinsky,
    title = "Kandinsky 3: Text-to-Image Synthesis for Multifunctional Generative Framework",
    author = "Vladimir, Arkhipkin  and
      Vasilev, Viacheslav  and
      Filatov, Andrei  and
      Pavlov, Igor  and
      Agafonova, Julia  and
      Gerasimenko, Nikolai  and
      Averchenkova, Anna  and
      Mironova, Evelina  and
      Anton, Bukashkin  and
      Kulikov, Konstantin  and
      Kuznetsov, Andrey  and
      Dimitrov, Denis",
    editor = "Hernandez Farias, Delia Irazu  and
      Hope, Tom  and
      Li, Manling",
    booktitle = "Proceedings of the 2024 Conference on Empirical Methods in Natural Language Processing: System Demonstrations",
    month = nov,
    year = "2024",
    address = "Miami, Florida, USA",
    publisher = "Association for Computational Linguistics",
    url = "https://aclanthology.org/2024.emnlp-demo.48/",
    doi = "10.18653/v1/2024.emnlp-demo.48",
    pages = "475--485"
}

@misc{kouzelis2025eqvaeequivarianceregularizedlatent,
      title={EQ-VAE: Equivariance Regularized Latent Space for Improved Generative Image Modeling}, 
      author={Theodoros Kouzelis and Ioannis Kakogeorgiou and Spyros Gidaris and Nikos Komodakis},
      year={2025},
      eprint={2502.09509},
      archivePrefix={arXiv},
      primaryClass={cs.LG},
      url={https://arxiv.org/abs/2502.09509}, 
}

@misc{skorokhodov2025improvingdiffusabilityautoencoders,
      title={Improving the Diffusability of Autoencoders}, 
      author={Ivan Skorokhodov and Sharath Girish and Benran Hu and Willi Menapace and Yanyu Li and Rameen Abdal and Sergey Tulyakov and Aliaksandr Siarohin},
      year={2025},
      eprint={2502.14831},
      archivePrefix={arXiv},
      primaryClass={cs.CV},
      url={https://arxiv.org/abs/2502.14831}, 
}

@misc{Higgins2016betaVAELB,
  title={beta-VAE: Learning Basic Visual Concepts with a Constrained Variational Framework},
  author={Irina Higgins and Lo{\"i}c Matthey and Arka Pal and Christopher P. Burgess and Xavier Glorot and Matthew M. Botvinick and Shakir Mohamed and Alexander Lerchner},
  booktitle={International Conference on Learning Representations},
  year={2016},
  url={https://api.semanticscholar.org/CorpusID:46798026}
}

@inproceedings{10.5555/3295222.3295378,
author = {van den Oord, Aaron and Vinyals, Oriol and Kavukcuoglu, Koray},
title = {Neural discrete representation learning},
year = {2017},
isbn = {9781510860964},
publisher = {Curran Associates Inc.},
address = {Red Hook, NY, USA},
booktitle = {Proceedings of the 31st International Conference on Neural Information Processing Systems},
pages = {6309–6318},
numpages = {10},
location = {Long Beach, California, USA},
series = {NIPS'17}
}

@misc{pmlr-v139-rybkin21a,
  title = 	 {Simple and Effective VAE Training with Calibrated Decoders},
  author =       {Rybkin, Oleh and Daniilidis, Kostas and Levine, Sergey},
  booktitle = 	 {Proceedings of the 38th International Conference on Machine Learning},
  pages = 	 {9179--9189},
  year = 	 {2021},
  editor = 	 {Meila, Marina and Zhang, Tong},
  volume = 	 {139},
  series = 	 {Proceedings of Machine Learning Research},
  month = 	 {07},
  publisher =    {PMLR},
  url = 	 {https://proceedings.mlr.press/v139/rybkin21a.html}
}

@misc{child2021deepvaesgeneralizeautoregressive,
      title={Very Deep VAEs Generalize Autoregressive Models and Can Outperform Them on Images}, 
      author={Rewon Child},
      year={2021},
      eprint={2011.10650},
      archivePrefix={arXiv},
      primaryClass={cs.LG},
      url={https://arxiv.org/abs/2011.10650}, 
}

@misc{jiang2021focal,
      title={Focal Frequency Loss for Image Reconstruction and Synthesis}, 
      author={Liming Jiang and Bo Dai and Wayne Wu and Chen Change Loy},
      year={2021},
      eprint={2012.12821},
      archivePrefix={arXiv},
      primaryClass={cs.CV},
      url={https://arxiv.org/abs/2012.12821}, 
}

@misc{pandey2022diffusevaeefficientcontrollablehighfidelity,
      title={DiffuseVAE: Efficient, Controllable and High-Fidelity Generation from Low-Dimensional Latents}, 
      author={Kushagra Pandey and Avideep Mukherjee and Piyush Rai and Abhishek Kumar},
      year={2022},
      eprint={2201.00308},
      archivePrefix={arXiv},
      primaryClass={cs.LG},
      url={https://arxiv.org/abs/2201.00308}, 
}

@misc{nvidia2025cosmosworldfoundationmodel,
      title={Cosmos World Foundation Model Platform for Physical AI}, 
      author={NVIDIA},
      year={2025},
      eprint={2501.03575},
      archivePrefix={arXiv},
      primaryClass={cs.CV},
      url={https://arxiv.org/abs/2501.03575}, 
}

@misc{flux2024,
    author={Black Forest Labs},
    title={FLUX},
    year={2024},
    howpublished={\url{https://github.com/black-forest-labs/flux}},
}

@misc{audiovae,
      title={Fast Timing-Conditioned Latent Audio Diffusion}, 
      author={Zach Evans and CJ Carr and Josiah Taylor and Scott H. Hawley and Jordi Pons},
      year={2024},
      eprint={2402.04825},
      archivePrefix={arXiv},
      primaryClass={cs.SD},
      url={https://arxiv.org/abs/2402.04825}, 
}

@misc{tolstikhin2019wassersteinautoencoders,
      title={Wasserstein Auto-Encoders}, 
      author={Ilya Tolstikhin and Olivier Bousquet and Sylvain Gelly and Bernhard Schoelkopf},
      year={2019},
      eprint={1711.01558},
      archivePrefix={arXiv},
      primaryClass={stat.ML},
      url={https://arxiv.org/abs/1711.01558}, 
}

@misc{dilokthanakul2017deepunsupervisedclusteringgaussian,
      title={Deep Unsupervised Clustering with Gaussian Mixture Variational Autoencoders}, 
      author={Nat Dilokthanakul and Pedro A. M. Mediano and Marta Garnelo and Matthew C. H. Lee and Hugh Salimbeni and Kai Arulkumaran and Murray Shanahan},
      year={2017},
      eprint={1611.02648},
      archivePrefix={arXiv},
      primaryClass={cs.LG},
      url={https://arxiv.org/abs/1611.02648}, 
}

@misc{10.1609/aaai.v33i01.33015066,
    author = {Takahashi, Hiroshi and Iwata, Tomoharu and Yamanaka, Yuki and Yamada, Masanori and Yagi, Satoshi},
    title = {Variational autoencoder with implicit optimal priors},
    year = {2019},
    isbn = {978-1-57735-809-1},
    publisher = {AAAI Press},
    url = {https://doi.org/10.1609/aaai.v33i01.33015066},
    doi = {10.1609/aaai.v33i01.33015066},
    booktitle = {Proceedings of the Thirty-Third AAAI Conference on Artificial Intelligence and Thirty-First Innovative Applications of Artificial Intelligence Conference and Ninth AAAI Symposium on Educational Advances in Artificial Intelligence},
    articleno = {622},
    numpages = {8},
    location = {Honolulu, Hawaii, USA},
    series = {AAAI'19/IAAI'19/EAAI'19}
}

@inproceedings{10.5555/3540261.3541252,
author = {Sinha, Samarth and Dieng, Adji B.},
title = {Consistency regularization for variational auto-encoders},
year = {2021},
isbn = {9781713845393},
publisher = {Curran Associates Inc.},
address = {Red Hook, NY, USA},
booktitle = {Proceedings of the 35th International Conference on Neural Information Processing Systems},
articleno = {991},
numpages = {12},
series = {NIPS '21}
}

@misc{pmlr-v80-zhao18b,
      title={Adversarially Regularized Autoencoders}, 
      author={Jake Zhao and Yoon Kim and Kelly Zhang and Alexander M. Rush and Yann LeCun},
      year={2018},
      eprint={1706.04223},
      archivePrefix={arXiv},
      primaryClass={cs.LG},
      url={https://arxiv.org/abs/1706.04223}, 
}

@misc{ramesh2022hierarchicaltextconditionalimagegeneration,
      title={Hierarchical Text-Conditional Image Generation with CLIP Latents}, 
      author={Aditya Ramesh and Prafulla Dhariwal and Alex Nichol and Casey Chu and Mark Chen},
      year={2022},
      eprint={2204.06125},
      archivePrefix={arXiv},
      primaryClass={cs.CV},
      url={https://arxiv.org/abs/2204.06125}, 
}

@misc{yang2025cogvideoxtexttovideodiffusionmodels,
      title={CogVideoX: Text-to-Video Diffusion Models with An Expert Transformer}, 
      author={Zhuoyi Yang and Jiayan Teng and Wendi Zheng and Ming Ding and Shiyu Huang and Jiazheng Xu and Yuanming Yang and Wenyi Hong and Xiaohan Zhang and Guanyu Feng and Da Yin and Yuxuan Zhang and Weihan Wang and Yean Cheng and Bin Xu and Xiaotao Gu and Yuxiao Dong and Jie Tang},
      year={2025},
      eprint={2408.06072},
      archivePrefix={arXiv},
      primaryClass={cs.CV},
      url={https://arxiv.org/abs/2408.06072}, 
}

@misc{kong2025hunyuanvideosystematicframeworklarge,
      title={HunyuanVideo: A Systematic Framework For Large Video Generative Models}, 
      author={Weijie Kong and Qi Tian and Zijian Zhang and Rox Min and Zuozhuo Dai and Jin Zhou and Jiangfeng Xiong and Xin Li and Bo Wu and Jianwei Zhang and Kathrina Wu and Qin Lin and Junkun Yuan and Yanxin Long and Aladdin Wang and Andong Wang and Changlin Li and Duojun Huang and Fang Yang and Hao Tan and Hongmei Wang and Jacob Song and Jiawang Bai and Jianbing Wu and Jinbao Xue and Joey Wang and Kai Wang and Mengyang Liu and Pengyu Li and Shuai Li and Weiyan Wang and Wenqing Yu and Xinchi Deng and Yang Li and Yi Chen and Yutao Cui and Yuanbo Peng and Zhentao Yu and Zhiyu He and Zhiyong Xu and Zixiang Zhou and Zunnan Xu and Yangyu Tao and Qinglin Lu and Songtao Liu and Dax Zhou and Hongfa Wang and Yong Yang and Di Wang and Yuhong Liu and Jie Jiang and Caesar Zhong},
      year={2025},
      eprint={2412.03603},
      archivePrefix={arXiv},
      primaryClass={cs.CV},
      url={https://arxiv.org/abs/2412.03603}, 
}

@misc{blattmann2023videoldm,
    title={Align your Latents: High-Resolution Video Synthesis with Latent Diffusion Models},
    author={Blattmann, Andreas and Rombach, Robin and Ling, Huan and Dockhorn, Tim and Kim, Seung Wook and Fidler, Sanja and Kreis, Karsten},
    booktitle={IEEE Conference on Computer Vision and Pattern Recognition ({CVPR})},
    year={2023},
}

@article{Kingma_2019,
   title={An Introduction to Variational Autoencoders},
   volume={12},
   ISSN={1935-8245},
   url={http://dx.doi.org/10.1561/2200000056},
   DOI={10.1561/2200000056},
   number={4},
   journal={Foundations and Trends® in Machine Learning},
   publisher={Now Publishers},
   author={Kingma, Diederik P. and Welling, Max},
   year={2019},
   pages={307–392} 
}

@InProceedings{10.1007/978-3-031-72998-0_17,
    author="Tschannen, Michael
    and Eastwood, Cian
    and Mentzer, Fabian",
    editor="Leonardis, Ale{\v{s}}
    and Ricci, Elisa
    and Roth, Stefan
    and Russakovsky, Olga
    and Sattler, Torsten
    and Varol, G{\"u}l",
    title="GIVT: Generative Infinite-Vocabulary Transformers",
    booktitle="Computer Vision -- ECCV 2024",
    year="2025",
    publisher="Springer Nature Switzerland",
    address="Cham",
    pages="292--309",
    isbn="978-3-031-72998-0"
}

@ARTICLE{10815947,
  author={Arkhipkin, Vladimir and Shaheen, Zein and Vasilev, Viacheslav and Dakhova, Elizaveta and Sobolev, Konstantin and Kuznetsov, Andrey and Dimitrov, Denis},
  journal={IEEE Access}, 
  title={ImproveYourVideos: Architectural Improvements for Text-to-Video Generation Pipeline}, 
  year={2025},
  volume={13},
  number={},
  pages={1986-2003},
  doi={10.1109/ACCESS.2024.3522510}}

@misc{Hessel2021CLIPScoreAR,
  title={CLIPScore: A Reference-free Evaluation Metric for Image Captioning},
  author={Jack Hessel and Ari Holtzman and Maxwell Forbes and Ronan Le Bras and Yejin Choi},
  journal={ArXiv},
  year={2021},
  volume={abs/2104.08718},
  url={https://api.semanticscholar.org/CorpusID:233296711}
}

@article{schuhmann2022laion,
  title={Laion-5b: An open large-scale dataset for training next generation image-text models},
  author={Schuhmann, Christoph and Beaumont, Romain and Vencu, Richard and Gordon, Cade and Wightman, Ross and Cherti, Mehdi and Coombes, Theo and Katta, Aarush and Mullis, Clayton and Wortsman, Mitchell and others},
  journal={Advances in neural information processing systems},
  volume={35},
  pages={25278--25294},
  year={2022}
}

@misc{bai2025qwen2,
  title={Qwen2. 5-vl technical report},
  author={Bai, Shuai and Chen, Keqin and Liu, Xuejing and Wang, Jialin and Ge, Wenbin and Song, Sibo and Dang, Kai and Wang, Peng and Wang, Shijie and Tang, Jun and others},
  journal={arXiv preprint arXiv:2502.13923},
  year={2025}
}

@misc{radford2021learning,
  title={Learning transferable visual models from natural language supervision},
  author={Radford, Alec and Kim, Jong Wook and Hallacy, Chris and Ramesh, Aditya and Goh, Gabriel and Agarwal, Sandhini and Sastry, Girish and Askell, Amanda and Mishkin, Pamela and Clark, Jack and others},
  booktitle={International conference on machine learning},
  pages={8748--8763},
  year={2021},
  organization={PmLR}
}

@misc{ma2024exploring,
  title={Exploring the role of large language models in prompt encoding for diffusion models},
  author={Ma, Bingqi and Zong, Zhuofan and Song, Guanglu and Li, Hongsheng and Liu, Yu},
  journal={arXiv preprint arXiv:2406.11831},
  year={2024}
}

@misc{Han2013,
  title={Comparison of Commonly Used Image Interpolation Methods},
  author={Dianyuan Han},
  year={2013},
  booktitle={Proceedings of the 2nd International Conference on Computer Science and Electronics Engineering (ICCSEE 2013)},
  pages={1556-1559},
  issn={1951-6851},
  isbn={978-90-78677-61-1},
  url={https://doi.org/10.2991/iccsee.2013.391},
  doi={10.2991/iccsee.2013.391},
  publisher={Atlantis Press}
}

@misc{Bhatt,
    author = {Bhatt, Parth and Pandit, Rakesh},
    year = {2013},
    month = {01},
    pages = {},
    title = {Comparative Analysis of Interpolation and Texture Synthesis Method for Enhancing Image}
}

@misc{wu2018groupnormalization,
      title={Group Normalization}, 
      author={Yuxin Wu and Kaiming He},
      year={2018},
      eprint={1803.08494},
      archivePrefix={arXiv},
      primaryClass={cs.CV},
      url={https://arxiv.org/abs/1803.08494}, 
}

@misc{5206848,
  author={Deng, Jia and Dong, Wei and Socher, Richard and Li, Li-Jia and Kai Li and Li Fei-Fei},
  booktitle={2009 IEEE Conference on Computer Vision and Pattern Recognition}, 
  title={ImageNet: A large-scale hierarchical image database}, 
  year={2009},
  pages={248-255},
  doi={10.1109/CVPR.2009.5206848}
}

@misc{lin2015microsoftcococommonobjects,
      title={Microsoft COCO: Common Objects in Context}, 
      author={Tsung-Yi Lin and Michael Maire and Serge Belongie and Lubomir Bourdev and Ross Girshick and James Hays and Pietro Perona and Deva Ramanan and C. Lawrence Zitnick and Piotr Dollár},
      year={2015},
      eprint={1405.0312},
      archivePrefix={arXiv},
      primaryClass={cs.CV},
      url={https://arxiv.org/abs/1405.0312}, 
}

@misc{karras2019stylebasedgeneratorarchitecturegenerative,
      title={A Style-Based Generator Architecture for Generative Adversarial Networks}, 
      author={Tero Karras and Samuli Laine and Timo Aila},
      year={2019},
      eprint={1812.04948},
      archivePrefix={arXiv},
      primaryClass={cs.NE},
      url={https://arxiv.org/abs/1812.04948}, 
}

@misc{xu2019understandingimprovinglayernormalization,
      title={Understanding and Improving Layer Normalization}, 
      author={Jingjing Xu and Xu Sun and Zhiyuan Zhang and Guangxiang Zhao and Junyang Lin},
      year={2019},
      eprint={1911.07013},
      archivePrefix={arXiv},
      primaryClass={cs.LG},
      url={https://arxiv.org/abs/1911.07013}, 
}

\appendix

\newpage
\section{Implementation Details}

\subsection{VAE Blocks}\label{appendix:blocks}

\begin{figure*}[!ht]
  \centering
  \includegraphics[bb=0 0 1869 812, width=0.75\textwidth]{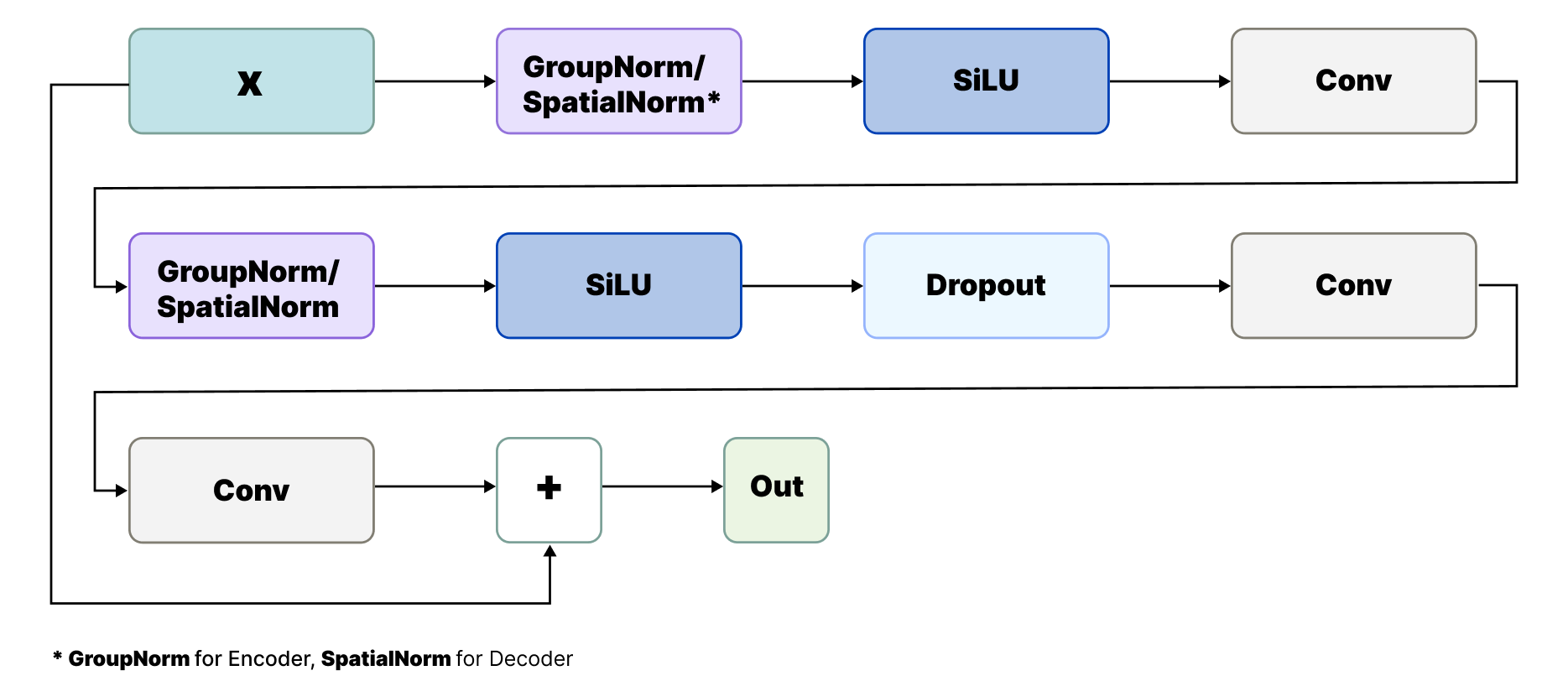}
  \caption{The ResNet Block for the VAE Encoder or Decoder. We use the Group Normalization \cite{wu2018groupnormalization} for the Encoder and Spatially Conditional Normalization \cite{zheng2022movq} for the Decoder.}
  \label{fig:resnet}
\end{figure*}

\begin{figure*}[!h]
  \centering
  \includegraphics[bb=0 0 3127 594, width=\textwidth]{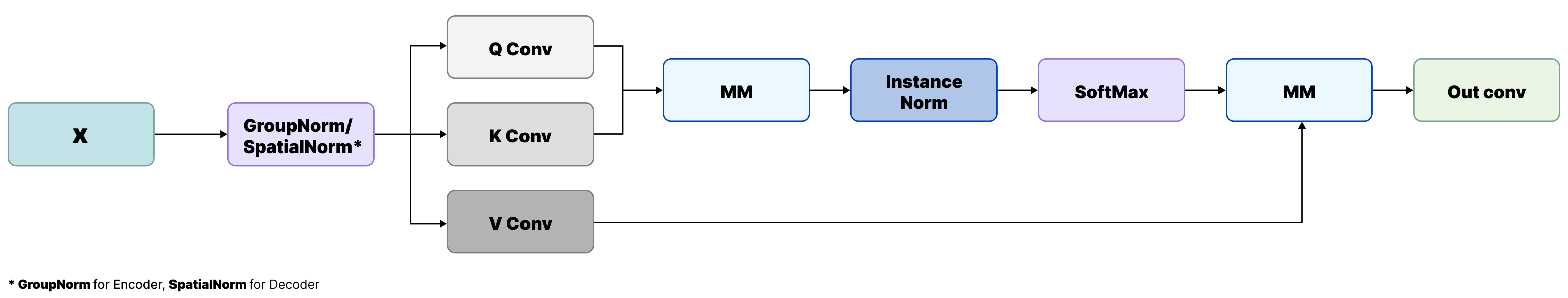}
  \caption{The Self-Attention Block for the VAE Encoder or Decoder. We use the Group Normalization \cite{wu2018groupnormalization} for the Encoder and Spatially Conditional Normalization \cite{zheng2022movq} for the Decoder.}
  \label{fig:attention}
\end{figure*}

\subsection{VAE Training}\label{appendix:vae}
Our variational autoencoder was implemented with the configuration presented in the Table \ref{tab:vae_training}.

\begin{table}[h]
\centering
\caption{VAE training parameters}
\begin{tabular}{ll}
\toprule
Parameter & Value \\
\midrule
Precision & FP32 \\
Learning rate ($\eta$) & $1 \times 10^{-4}$ \\
EMA decay ($\gamma$) & 0.9999 \\
Latent dimension ($d_z$) & 16 \\
Discriminator start & Step 0 \\
\bottomrule
\end{tabular}
\label{tab:vae_training}
\end{table}

The optimal loss weights are presented in the Table \ref{tab:vae_losses}.

\begin{table}[!h]
\centering
\caption{VAE losses coefficients}
\begin{tabular}{lc}
\toprule
Component & Weight ($\lambda$) \\
\midrule
KL Divergence Loss & $1 \times 10^{-4}$ \\
$L_2$ Reconstruction Loss & 1.0 \\
Adversarial Loss & 0.01 \\
Perceptual Loss & 0.1 \\
\bottomrule
\end{tabular}
\label{tab:vae_losses}
\end{table}

\subsection{Diffusion Transformer (DiT)}\label{appendix:dit}
The 2 billion parameter diffusion transformer employed the following architecture:

\subsubsection{Embedding Systems}
\begin{itemize}
\item \textbf{Qwen 2.5 VL 7B Instruct} \cite{bai2025qwen2}:
\begin{itemize}
\item Embedding dimension: $d_{\text{Qwen}} = 3584$
\item Context length: $L_{\text{Qwen}} = 256$
\end{itemize}

\item \textbf{CLIP} \cite{radford2021learning}:
\begin{itemize}
\item Embedding dimension: $d_{\text{CLIP}} = 768$
\item Context length: $L_{\text{CLIP}} = 77$
\end{itemize}
\end{itemize}

\subsubsection{Model Architecture}
\begin{itemize}
\item Latent patch size: $2 \times 2$
\item Time embedding dimension: $d_t = 512$
\item Hidden dimension: $d_h = 1792$
\item Feed-forward dimension: $d_{ff} = 7168$
\item Transformer blocks: $N = 32$
\end{itemize}

\subsubsection{Optimization Parameters}
\begin{table}[h]
\centering
\caption{Training Configuration}
\begin{tabular}{ll}
\toprule
Parameter & Value \\
\midrule
Optimizer & AdamW \\
Learning rate ($\eta$) & $1 \times 10^{-4}$ \\
Gradient clipping & $\|\nabla\|_2 \leq 1.0$ \\
Weight decay & 0.0 \\
$\beta_1$, $\beta_2$ & 0.9, 0.95 \\
$\epsilon$ & $1 \times 10^{-8}$ \\
LR schedule & Constant + 1000 warmup \\
\bottomrule
\end{tabular}
\end{table}

\newpage
\section{Additional Results on Artifact Mitigation}\label{appendix:results}

\begin{figure*}[!ht]
  \centering
  \includegraphics[bb=0 0 1726 1053, width=\textwidth]{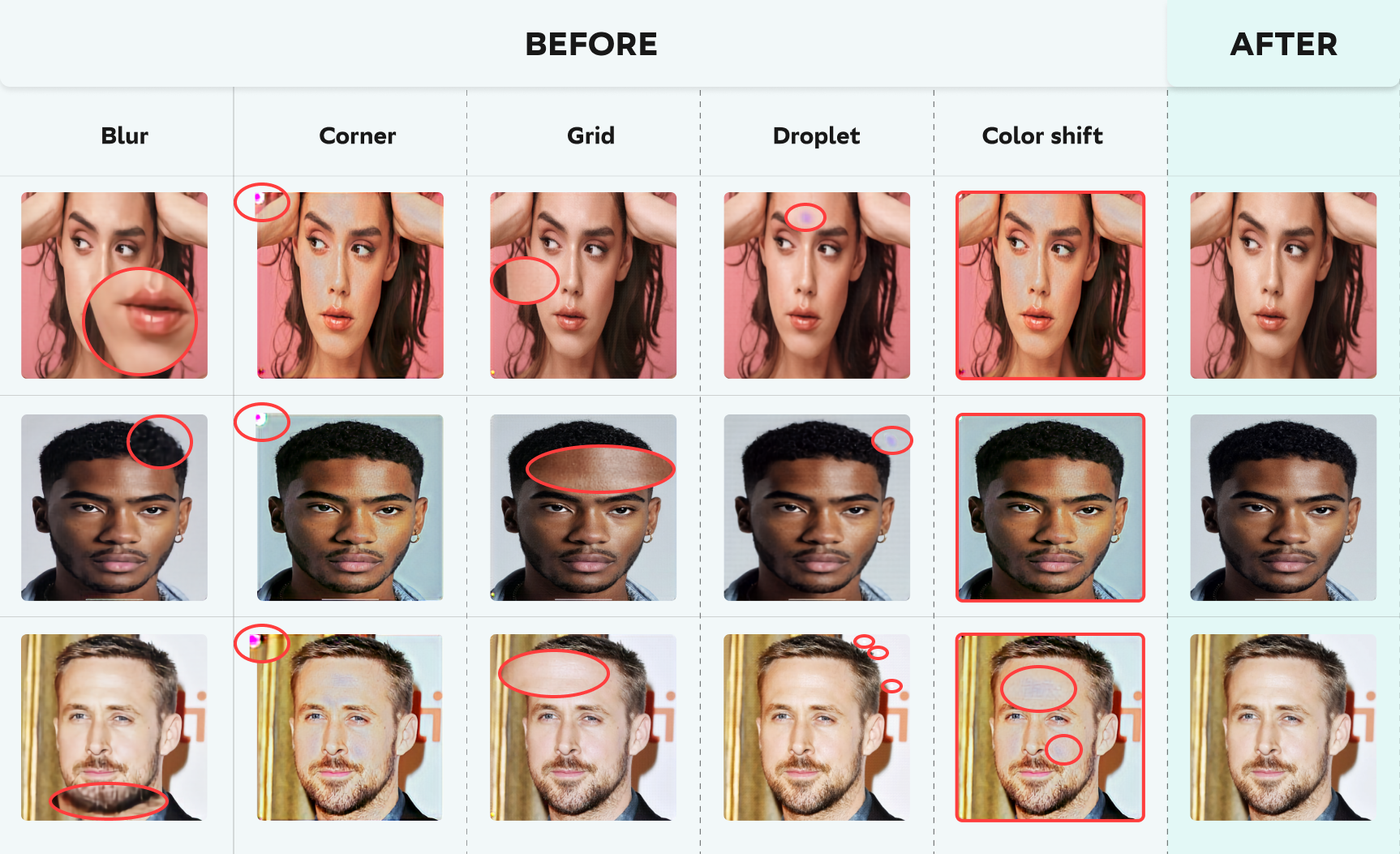}
  \label{fig:artifacts_mitigation_appendix}
\end{figure*}

\begin{figure*}[!h]
  \centering
  \includegraphics[bb=0 0 1726 780, width=\textwidth]{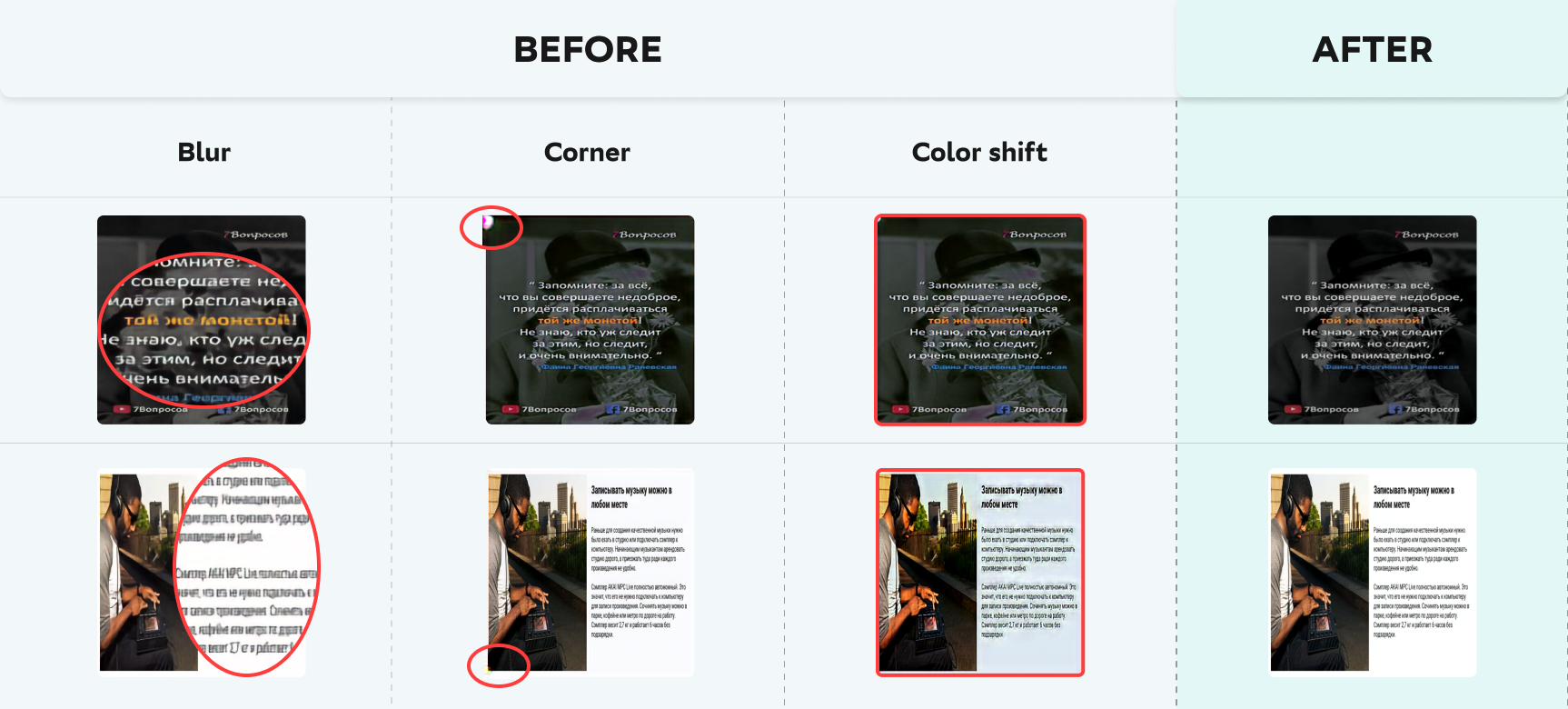}
  \label{fig:artifacts_mitigation_appendix_2}
\end{figure*}

\end{document}